\documentclass[journal,twoside,web]{ieeecolor}
\usepackage{jsen}
\usepackage{cite}
\usepackage{amsmath,amssymb,amsfonts}
\usepackage{algorithmic}

\usepackage{textcomp}
\usepackage{wrapfig}

\usepackage{amsmath,amsfonts}
\usepackage{array}
\usepackage{textcomp}
\usepackage{stfloats}
\usepackage{url}
\usepackage{verbatim}
\usepackage{cite}
\usepackage{amssymb}

\usepackage[ruled,vlined]{algorithm2e}
\usepackage{hyperref}
\usepackage{float}
\usepackage{multirow}
\usepackage{booktabs,caption}
\usepackage[flushleft]{threeparttable}
\usepackage{subcaption}
\floatstyle{plaintop}
\restylefloat{table}

\usepackage{stackengine}
\usepackage{arydshln}
\usepackage[pdftex]{graphicx}

\usepackage{fancyhdr}
\usepackage{kantlipsum}
\usepackage{eso-pic}

\def\BibTeX{{\rm B\kern-.05em{\sc i\kern-.025em b}\kern-.08em
    T\kern-.1667em\lower.7ex\hbox{E}\kern-.125emX}}
\definecolor{abstractbg}{rgb}{0.89804,0.94510,0.83137}
\usepackage{soul} % For highlighting
\usepackage{color} % For color definitions
\DeclareRobustCommand{\rev}[1]{{\sethlcolor{yellow}\hl{#1}}}

\DeclareRobustCommand{\revx}[1]{{\sethlcolor{green}\hl{#1}}}

\DeclareRobustCommand{\revz}[1]{{\sethlcolor{cyan}\hl{#1}}}

\DeclareRobustCommand{\rev}[1]{#1} % <--- UNCOMMENT THIS LINE LATER FOR CLEAN VERSION

\DeclareRobustCommand{\revx}[1]{#1} % <--- UNCOMMENT THIS LINE LATER FOR CLEAN VERSION

\DeclareRobustCommand{\revz}[1]{#1} % <--- UNCOMMENT THIS LINE LATER FOR CLEAN VERSION

\soulregister\cite7
\soulregister\ref7
\soulregister\textbf7
\begin{document}

\AddToShipoutPictureBG*{%
	\AtPageLowerLeft{%
		\setlength\unitlength{1in}%
		\hspace*{\dimexpr0.5\paperwidth\relax}%%  change \dimexpr0.5\paperwidth\relax appropriately
		\makebox(0,0.63)[c]{1558-1748~\copyright2026 IEEE. This work has been accepted for publication in IEEE Sensors Journal.}
		\makebox(0,0.3)[c]{The published version can be accessed at \href{https://ieeexplore.ieee.org/document/11373257/}{https://ieeexplore.ieee.org/document/11373257/}. DOI: \href{https://doi.org/10.1109/JSEN.2026.3656442}{10.1109/JSEN.2026.3656442}}
}}

\title{Seq-DeepIPC: Sequential Sensing for End-to-End Control in Legged Robot Navigation}

\author{Oskar Natan, \IEEEmembership{Member, IEEE}, and Jun Miura, \IEEEmembership{Member, IEEE}
\thanks{Oskar Natan (corresponding author) is with the Department of Computer Science and Electronics, Universitas Gadjah Mada, Yogyakarta Indonesia, Email: oskarnatan@ugm.ac.id}
\thanks{Jun Miura is with the Department of Computer Science and Engineering, Toyohashi University of Technology, Toyohashi Japan. Email: jun.miura@tut.jp}}

\IEEEtitleabstractindextext{%
\fcolorbox{abstractbg}{abstractbg}{%
\begin{minipage}{\textwidth}%

%\begin{wrapfigure}[12]{r}{3in}
%	\centering
%	\resizebox{0.98\linewidth}{!}{%
%		\begin{tikzpicture}[
%			node distance=4.5mm, >=Latex, semithick,
%			box/.style={draw, rounded corners, align=center,
%				minimum width=1.55cm, minimum height=0.68cm, font=\scriptsize},
%			every node/.style={inner sep=1pt}
%			]
%			% Nodes (short labels)
%			\node[box, fill=blue!10]                 (in)   {Seq. RGB--D};
%			\node[box, fill=green!10, right=of in]   (perc) {Perception\\(Seg+Depth)};
%			\node[box, fill=orange!15, below=of perc,yshift=1mm] (bev)  {BEV\\Projection};
%			\node[box, fill=yellow!30, right=of perc] (gru) {Fusion+GRU\\Planning};
%			\node[box, fill=red!10, right=of gru]     (nav) {Legged\\Navigation};
%			
%			% Arrows
%			\draw[->] (in) -- (perc);
%			\draw[->] (perc) -- (gru);
%			\draw[->] (bev) -- (gru);
%			\draw[->] (gru) -- (nav);
%			\draw[->, dashed] (perc) -- (bev);
%			
%			% Tiny annotations
%			\node[font=\scriptsize, above=0mm of perc] {Multi-task};
%			\node[font=\scriptsize, below=0mm of bev]  {BEV context};
%			\node[font=\scriptsize, above=0mm of gru]  {Temporal features};
%		\end{tikzpicture}%
%	}
%\end{wrapfigure}

\begin{wrapfigure}[11]{r}{3in}%
	\includegraphics[width=3in]{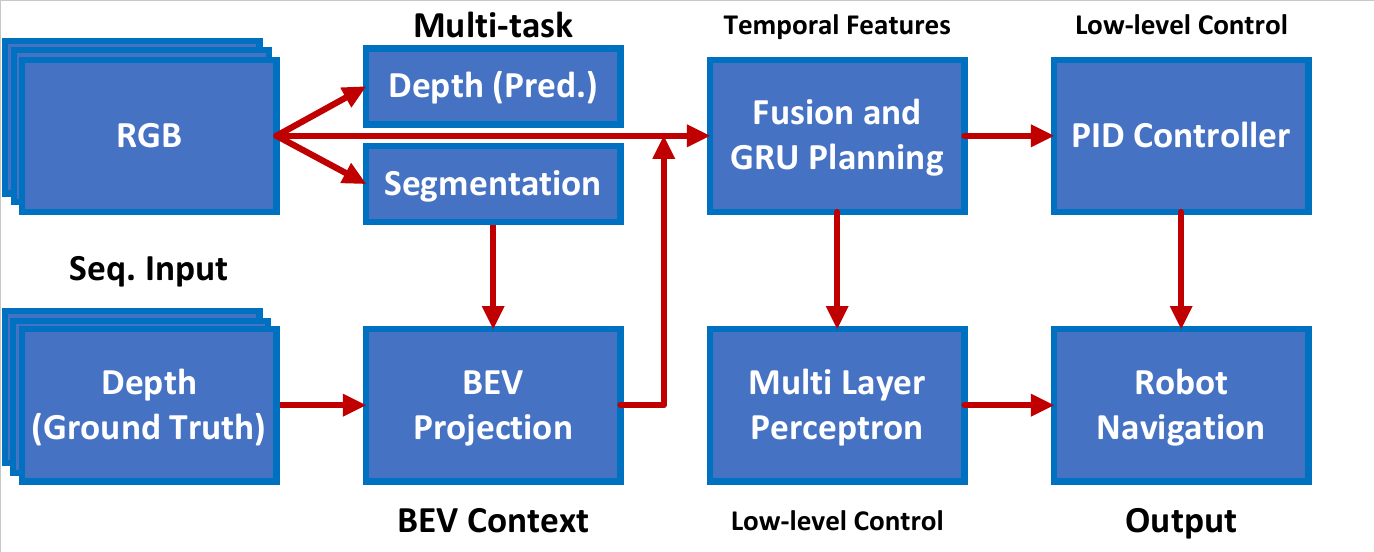}%
\end{wrapfigure}%

\begin{abstract}
	
We present Seq-DeepIPC, a sequential end-to-end perception-to-control model for legged robot navigation in real-world environments. Seq-DeepIPC advances intelligent sensing for autonomous legged navigation by tightly integrating multi-modal perception (RGB-D + GNSS) with temporal fusion and control. The model jointly predicts semantic segmentation and depth estimation, giving richer spatial features for planning and control. For efficient deployment on edge devices, we use a lightweight model as the encoder, reducing computation while maintaining accuracy. Heading estimation is simplified by removing the noisy IMU and instead deriving global heading via differential analysis of sequential GNSS coordinates. We collected a larger and more diverse dataset that includes both road and grass terrains, and validated Seq-DeepIPC on a robot dog. Comparative and ablation studies show that sequential inputs improve perception and control in our models, while other baselines do not benefit. Seq-DeepIPC achieves competitive or better results with reasonable model size; although GNSS-only heading is less reliable near tall buildings, it is robust in open areas. Overall, Seq-DeepIPC extends end-to-end navigation beyond wheeled robots to more versatile and temporally-aware systems. To support future research, we will release the codes to our GitHub repo at \href{https://github.com/oskarnatan/Seq-DeepIPC}{https://github.com/oskarnatan/Seq-DeepIPC}.
\end{abstract}

\end{minipage}}}

\maketitle

\section{Introduction}
%(e.g., separate perception, mapping, planning) 
Modern robotic navigation systems increasingly favor multi-modal end-to-end learning architectures, which directly map raw sensory data to control commands \cite{mimoend2end}. This approach reduces the need for hand-engineered modules and allows shared features and joint optimization across tasks \cite{Dingreview}\cite{chibreview}. Using this strategy, we can minimize information loss as the systems can learn all by itself. However, real-world deployment of such systems is still fraught with challenges: sensor noise, localization drift, dynamic environments, and variable terrain all undermine performance \cite{aizatnav}\cite{natanend}. Moreover, resource constraints on embedded platforms often limit how expressive such models can be. In many robotics settings, especially outdoors or on mobile agents, these issues degrade navigation performance and robustness \cite{wuend}\cite{dufuse}\cite{slamurban}. From a sensing perspective, the challenge lies in converting multi-modal, noisy sensor streams into consistent spatial representations that can guide control reliably in unstructured environments.

To address these challenges, recent methods have incorporated multi-task supervision, sensor fusion, and temporal modeling \cite{zhangmtl}\cite{mimooskar}\cite{peirgbd}\cite{natan2025deepipcv2}. For example, Huang \textit{et al.} fuse RGB and depth at early or late stages to enhance spatial understanding \cite{huang_model}, other model called AIM-MT \cite{aim_mt} encourages the network to learn auxiliary tasks (semantic segmentation and depth estimation) to improve its latent representations. Our previous work, DeepIPC, made strides by coupling segmentation-guided feature extraction with waypoint-driven control in a compact architecture, demonstrating real-world drivability on a wheeled platform \cite{natan2024deepipc}. However, three persistent gaps remain: (i) single-frame perception is susceptible to aliasing and temporal inconsistency; (ii) reliance on IMU-based heading estimation introduces drift, magnetic interference, and hardware complexity; (iii) most existing systems are validated on wheeled robots navigating structured roads, limiting generality to more challenging surfaces.% (e.g., grass, uneven terrain). , especially in low-texture or dynamic scenes

%\cite{Qionglight}\cite{Salehreal}
In this paper, we present Seq-DeepIPC as a continuation and improvement of DeepIPC \cite{natan2024deepipc}. Our model ingests short sequences of RGB-D frames, producing temporally consistent features, and jointly estimates semantic segmentation + depth to enable stronger spatial reasoning. We replace heavier encoders with EfficientNet-B0, facilitating real-time inference on constrained platforms \cite{tan2019efficientnet}\cite{Lammielow}. Rather than using a noisy IMU, we compute bearing angle from consecutive GNSS fixes via a geodesic formula, which improves heading stability in open terrain. \rev{We evaluate the system on a larger campus loop, under mixed road, grass, and uneven terrains, and deploy it on a legged robot to test its generalization beyond wheeled navigation.} In comparative and ablation studies (varying sequence lengths, comparing with Huang \textit{et al.} \cite{huang_model} and AIM-MT \cite{aim_mt} baselines), we demonstrate that only our DeepIPC-derived models benefit from temporal input, achieving consistent gains in perception and control. Our key novelties are:

%\begin{enumerate}
%	\item \rev{A multi-task perception module that integrates sequential RGB-D inputs with dual-head supervision (segmentation and depth). This enriches spatial feature extraction and ensures geometric consistency for BEV mapping.}
%	\item \rev{A stability-focused control architecture that leverages GRU-based temporal fusion to mitigate the camera motion blur inherent to legged locomotion, enabling smooth control without complex mechanical stabilization.}
%	\item Deployment and evaluation on a legged robot navigating on road and uneven terrain in a larger-scale environment.
%\end{enumerate}

\begin{enumerate}
	\item \revx{A \textbf{Locomotion-Aware Temporal Perception} framework that integrates sequential RGB-D inputs ($K=3$) via a GRU. Unlike standard video-based methods, our temporal window is empirically tuned to mitigate the specific high-frequency camera pitch oscillations inherent to legged robot gaits, stabilizing BEV projection without mechanical stabilization.}
	\item \revx{A \textbf{Magnetometer-Free Global Heading} estimator that derives bearing solely from differential GNSS fixes. We demonstrate that this approach eliminates the standard IMU-based compasses drift caused by hard/soft iron magnetic interference in urban environments.}
	\item \revx{A comprehensive evaluation on a \textbf{Legged Robot in Mixed Terrain} (stairs, grass, asphalt), demonstrating that our sequential fusion significantly outperforms single-frame baselines and standard fusion models in handling the chaotic camera motion of legged platforms.}
\end{enumerate}

\section{Related Works}

\subsection{End-to-End Learning for Robot Navigation}
End-to-end navigation learns a direct mapping from raw onboard sensors to control, bypassing modular stacks. Recent works have advanced imitation and reinforcement learning formulations, sensor fusion, and decoder design. Ishihara \textit{et al.} introduced an attention-based multi-task policy (AIM-MT) that couples perception heads with driving commands inside a conditional imitation framework \cite{Ishihara2021AIMMT}. Hou and Zhang showed that by adding safety information into the probabilistic graphical model(PGM) and learning it in conjunction with the reinforcement learning process can solve multiple driving tasks \cite{houzang}. For non-vision modalities, Wang \textit{et al.} demonstrated end-to-end navigation from raw LiDAR with robustness gains \cite{wanglidar}. Transformer-based fusion has also become prominent in integrating image and LiDAR features \cite{Chitta2023TransFuserTPAMI}. Decoder capacity and refinement have been explored, which stacks coarse-to-fine reasoning to stabilize planning under complex scenes \cite{Jia2023ThinkTwice}. These advances reduce hand-crafted intermediates and motivate our sequential and multi-task formulation. %with self-attention and achieves state-of-the-art closed-loop results

\subsection{Sequential and Temporal Modeling}
Temporal aggregation stabilizes perception and improves control under motion, occlusion, and noise. Attention-based fusion and recurrent encoders have been used to integrate multi-frame evidence for planning (e.g., TransFuser’s temporal fusion) \cite{Chitta2023TransFuserTPAMI}. Cross-modal temporal fusion strategies (e.g., CrossFuser) further refine multi-sensor features and improve downstream decision quality \cite{Wu2023CrossFuser}. Policy-level fusion (PolicyFuser) combines complementary policies to exploit temporal context and multi-modality in closed loop \cite{Huang2023PolicyFuser}. Beyond specific architectures, recent studies formalize multi-camera BEV temporal fusion and contextual representation for end-to-end planning \cite{Azam2024ContextE2E}. Other spatio-temporal pipelines such as DeepSTEP confirm that recurrent or transformer-based temporal integration can improve prediction robustness \cite{Huch2023DeepSTEP}. These directions motivate Seq-DeepIPC to fuse short RGB-D sequences via GRU, yielding temporally smoothed latent states that feed both waypoint and control heads.

\subsection{Legged Robot Navigation and Perception}

Legged platforms introduce mixed-terrain challenges and dynamic foothold constraints that differ fundamentally from wheeled robots. \rev{Early learning-based approaches focused on blind locomotion using proprioceptive Reinforcement Learning (RL) to handle challenging terrain \cite{lee2020learning}. However, blind policies struggle with discrete obstacles or steps. Thus, recent works incorporate exteroceptive perception, such as elevation mapping, to enable perceptive locomotion \cite{miki2022learning}. While robust, map-based methods can suffer from drift and latency. Alternatively, end-to-end vision-based frameworks have emerged, transformer-based models have been used to fuse proprioception and vision for rough terrain traversal \cite{yang2022learning}, and lightweight vision pipelines have been deployed on small-scale quadrupeds \cite{dengrobio}. More recently, semantic navigation systems like RDog \cite{Cai2024RDog} and ViTAL \cite{Fahmi2022ViTAL} have demonstrated the importance of high-level scene understanding for foothold selection. Despite this progress, few works integrate semantic scene understanding, temporal consistency, and end-to-end control into a unified framework for large-scale outdoor navigation. Our Seq-DeepIPC bridges this gap by coupling sequential RGB-D perception with control, validating the approach on mixed terrains within a campus-scale environment.}

\section{Methodology}

\subsection{Problem Statement}

We study end-to-end navigation for a legged robot operating on mixed terrain. At each time $t$, the robot observes a short sequence of $K$ multimodal frames
\begin{equation}
	O_t = \{o_t, o_{t-1}, \ldots, o_{t-K+1}\}, \quad o_i = (I_i^R, I_i^D),
\end{equation}
where $I_i^R$ and $I_i^D$ are the RGB and depth images, respectively. Then, the task is conditioned on a sequence of route points $\mathcal{P}_t$ that prescribes the path from start to goal. Let
\begin{equation}
	\mathcal{P}_t = \left\{(\phi_t^{(m)}, \lambda_t^{(m)})\right\}_{m=1}^M
\end{equation}
denote the next $M$ route points expressed in global latitude–longitude coordinates. During execution, the upcoming two points ($M=2$) are converted into local Cartesian coordinates and used to guide the controller. The learning objective is to find parameters $\theta$ of a function
\begin{equation}
	f_\theta : (O_t, \mathcal{P}_t, g_t) \;\mapsto\; \{\hat{S}_t, \hat{D}_t, \hat{W}_t, u_t\},
\end{equation}
where $g_t$ is the current GNSS measurement, $\hat{S}_t$ and $\hat{D}_t$ are the predicted semantic and depth maps, $\hat{W}_t = \{\hat{w}_1, \ldots, \hat{w}_N\}$ are the $N=5$ future waypoints in the local BEV frame, and $u_t = (x, y, \theta)$ denotes the position–orientation control action. The future waypoints $\hat{W}_t$ are obtained from the robot’s ground-truth trajectory during data collection, where we consider up to five steps ahead for prediction. The control action $u_t$ corresponds to the recorded remote state that teleoperated the robot during data gathering. Training thus follows a supervised imitation learning paradigm: given a dataset
\begin{equation}
	\mathcal{D} = \left\{(O_t, \mathcal{P}_t, g_t; S_t, D_t, W_t, u_t)\right\}_{t=1}^T,
\end{equation}

\begin{figure*}[t]
	\centering
	\includegraphics[width=0.96\textwidth]{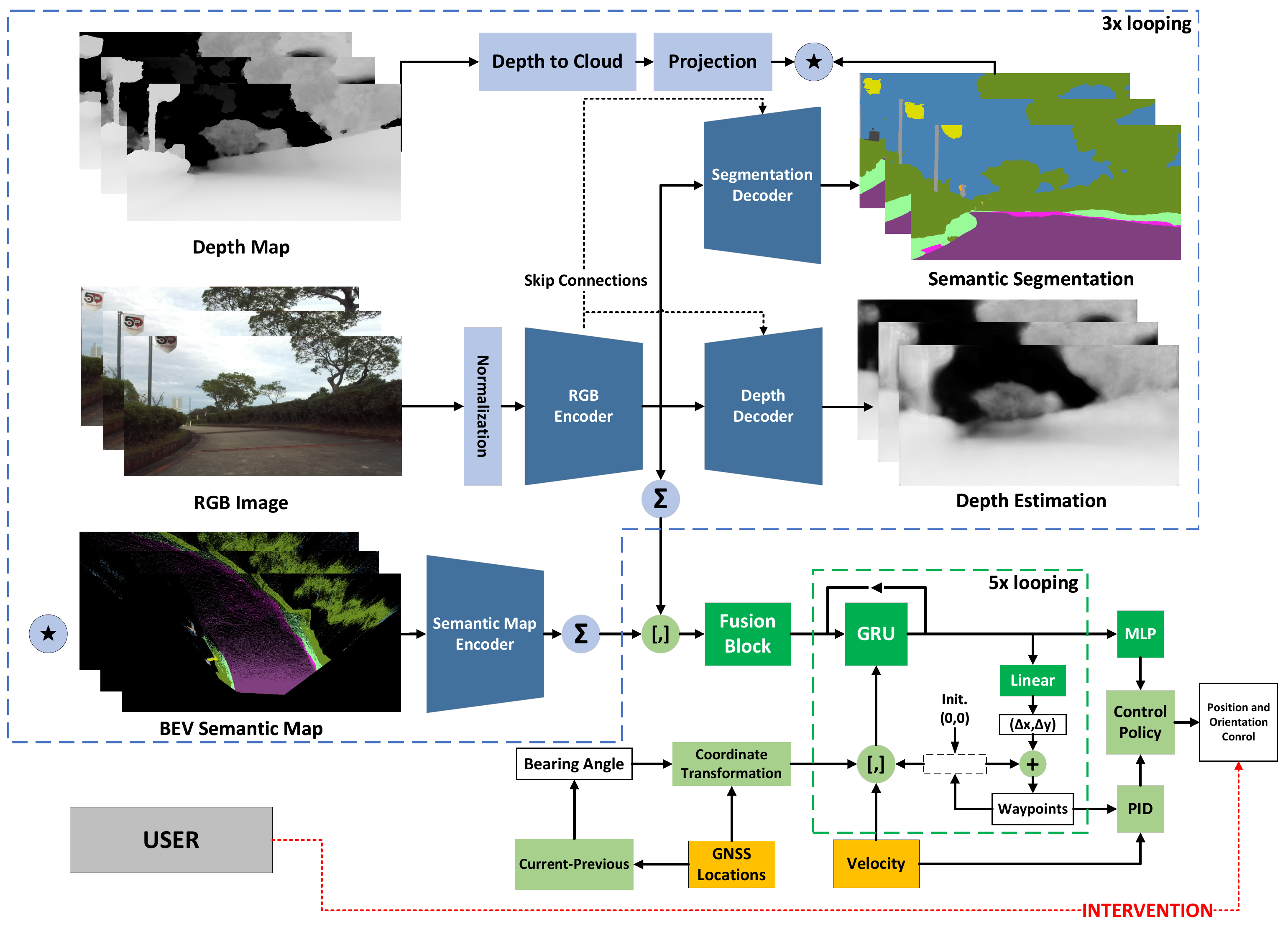}
	\caption{Overview of the proposed Seq-DeepIPC architecture.
		\revz{The framework is 'end-to-end' regarding the navigation policy (mapping raw sensor data to control commands). These commands are executed by the robot's internal built-in controller via inverse kinematics (see Subsection \ref{planandctrl}).}		
		(1) \textbf{Perception part}: Sequential RGB inputs are processed by an EfficientNet-B0 encoder, producing latent features that drive two prediction heads for semantic segmentation and depth estimation. The ground-truth depth maps are combined with predicted segmentation to generate BEV projections, which are further encoded by a second EfficientNet-B0 into BEV latent features. \revx{Notes: the Depth Map (Ground Truth, from the stereo camera sensing) is used to construct the target BEV map. Meanwhile, the Depth Estimation (predicted by the model) is used to supervise the RGB encoder in generating latent geometric features.}
		(2) \rev {\textbf{Planning and Control part}: Let $\mathbf{z}_t$ be the fused RGB and BEV latent features at time $t$, concatenated with transformed route points, bearing angle, and velocity are processed by a GRU to capture temporal dependencies. The resulting features drive two complementary control pathways: (a) PID controllers, which use predicted waypoints to estimate control signals, and (b) command-specific MLP controllers, which directly map the GRU latent space to $(x,y,\theta)$ controls. The blended control policy regulates position and orientation for the legged robot.}}
	\label{fig:arch}
\end{figure*}

we minimize the empirical risk
\begin{equation}
	\min_\theta \; \frac{1}{T}\sum_{t=1}^T \mathcal{L}_{\text{total}}\!\left(\hat{S}_t, \hat{D}_t, \hat{W}_t, u_t;\; S_t, D_t, W_t, u_t\right),
\end{equation}
where $\mathcal{L}_{\text{total}}$ aggregates the task-specific losses. All outputs $\{\hat{S}_t, \hat{D}_t, \hat{W}_t, u_t\}$ are defined in the local BEV frame whose origin is fixed at the robot base $(0,0)$.

\subsection{Proposed Model}
Seq-DeepIPC consists of two main components: (i) a \emph{perception part} that processes a short sequence of RGB-D observations and forms a BEV representation, and (ii) a \emph{planning \& control part} that infers future waypoints and control commands. An overview is shown in Fig.~\ref{fig:arch}.

\subsubsection{Perception Part}

At time $t$, the perception branch receives a sequence of $K$ RGB frames
\begin{equation}
	O_t = \{o_t, o_{t-1}, \ldots, o_{t-K+1}\}, \quad o_k = I^R_k,
\end{equation}
where $I^R_k$ denotes the $k$-th RGB image. \rev{Each frame is passed through a lightweight EfficientNet-B0 \cite{tan2019efficientnet} encoder to extract latent features $f^R_k$.} From this shared latent space, two prediction heads output a semantic segmentation map $\hat{S}_k$ that consists of 19 different object classes, as in the Cityscapes dataset \cite{Cordts2016Cityscapes}, and a depth map $\hat{D}_k$ that shows the distance on each pixel relative to the camera. While the predicted depth is not directly used for BEV construction, it serves as an auxiliary regression task that regularizes the encoder and enforces geometry-aware latent features.

\textbf{BEV construction.}
Let $\hat{S}_k\in[0,1]^{H_{\!img}\times W_{\!img}\times C}$ be the per-pixel class
scores and $I^D_k\in\mathbb{R}^{H_{\!img}\times W_{\!img}}$ the ground-truth depth.
Back-project pixels with intrinsics $\mathbf{K}$ and transform to the robot frame:
\begin{equation}
	\mathbf{X}^{r}=\mathcal{T}_{c\!\to r}\!\left(\pi^{-1}(\mathbf{K},\,I^D_k)\right)
	=\{(x_i,y_i,z_i)\}_i .
	\label{eq:bev_backproj}
\end{equation}

The BEV covers $x\in(0,16]$\,m (forward) and $y\in[-16,16]$\,m (left–right), with
resolution $H\times W \!=\! 128\times 256$ so that
$\Delta_x=\tfrac{16}{128}=0.125$\,m and $\Delta_y=\tfrac{32}{256}=0.125$\,m.
Index each point to the BEV grid (robot at bottom–center):
\begin{equation}
	\begin{aligned}
		i &= \big\lfloor x_i / \Delta_x \big\rfloor, \qquad
		j = \big\lfloor (y_i + 16) / \Delta_y \big\rfloor, \\
		(i,j) &\in [0,H{-}1] \times [0,W{-}1].
	\end{aligned}
	\label{eq:bev_index}
\end{equation}

Let $\mathbf{e}_{c}\in\{0,1\}^{C}$ be the one-hot vector for class $c$ and
$c_i=\arg\max_{c}\hat{S}_k^{(c)}(u_i,v_i)$. The BEV tensor
$M^{\mathrm{BEV}}_{k}\in\{0,1\}^{H\times W\times C}$ (with $C{=}20$ channels, one per
class) is obtained by “splatting’’ points to cells with a per-cell reducer: %(e.g., max/majority)
\begin{equation}
	M^{\mathrm{BEV}}_{k}(i,j,:) \leftarrow
	\mathrm{Agg}\big\{\,\mathbf{e}_{c_i}\;\big|\;(i,j)\text{ from \eqref{eq:bev_index}}\big\}.
	\label{eq:bev_agg}
\end{equation}

For sequential inputs, BEV maps are temporally fused by exponential smoothing:
\begin{equation}
	\tilde{M}^{\mathrm{BEV}}_{t}=\alpha\,M^{\mathrm{BEV}}_{t}
	+(1-\alpha)\,\tilde{M}^{\mathrm{BEV}}_{t-1},\quad \alpha\in(0,1],
	\label{eq:bev_ema}
\end{equation}
\noindent where the fused BEV tensor $\tilde{M}^{\mathrm{BEV}}_{t}\in\mathbb{R}^{128\times 256\times 20}$
is then encoded by EfficientNet-B0 to yield $f^{\mathrm{BEV}}_{t}$, which is fused with the RGB latent $f^{R}_{t}$ for planning and control. Importantly, the BEV map $\tilde{M}^{\mathsf{BEV}}_t$ is further encoded by a second EfficientNet-B0 encoder, producing a compact latent feature $f^{\text{BEV}}_t$. This BEV feature is then fused with the RGB encoder latent feature $f^R_t$ in the fusion block, forming the joint perception embedding $\mathbf{z}_t$ (front and BEV perspectives) that drives the planning and control module.

For simplicity, we do not perform ego-motion warping between frames; instead, the exponential decay term implicitly stabilizes the fused representation. The BEV generation assumes a locally planar ground, which is reasonable for most road and short-grass regions in our dataset. For strongly uneven or sloped terrain, this assumption introduces minor distortions, but these are mitigated by the temporal fusion that averages depth over multiple frames. In future work, incorporating pose-based spatial alignment and height-aware BEV encoding could further improve consistency in high-speed or non-planar motion.

\subsubsection{Planning and Control Part} \label{planandctrl}
Let $\mathbf{z}_t$ be the perception latent at time $t$ concatenated with ancillary inputs (e.g., robot speed) and the \emph{transformed} upcoming route points. A GRU models temporal dependencies:
\begin{align}
	\mathbf{r}_t &= \sigma\!\left(\mathbf{W}_r[\mathbf{z}_t,\mathbf{h}_{t-1}] + \mathbf{b}_r\right),\\
	\mathbf{u}_t &= \sigma\!\left(\mathbf{W}_u[\mathbf{z}_t,\mathbf{h}_{t-1}] + \mathbf{b}_u\right),\\
	\tilde{\mathbf{h}}_t &= \tanh\!\left(\mathbf{W}_h[\mathbf{z}_t,\mathbf{r}_t\odot\mathbf{h}_{t-1}] + \mathbf{b}_h\right),\\
	\mathbf{h}_t &= (1-\mathbf{u}_t)\odot \mathbf{h}_{t-1} + \mathbf{u}_t\odot \tilde{\mathbf{h}}_t,
\end{align}
where $\sigma(\cdot)$ is the logistic function and $\odot$ is the Hadamard product.

\textbf{PID controllers from waypoints.}
From $\mathbf{h}_t$ we predict incremental displacements $\Delta \mathbf{w}_\ell=(\Delta x_\ell,\Delta y_\ell)$ and roll them for $\ell=1{:}N$ steps:
\begin{align}
	\Delta \mathbf{w}_\ell &= \mathbf{W}^{(\ell)} \mathbf{h}_t + \mathbf{b}^{(\ell)},\qquad \ell=1,\ldots,N,\\
	\hat{\mathbf{w}}_\ell &= \hat{\mathbf{w}}_{\ell-1}+\Delta \mathbf{w}_\ell,\qquad \hat{\mathbf{w}}_0=(0,0),
\end{align}

\noindent where waypoints $\{\hat{\mathbf{w}}_\ell\}_{\ell=1}^N$ live in the local BEV frame. Then, we form an \emph{aim point} $\mathbf{a}=(\hat{\mathbf{w}}_1+\hat{\mathbf{w}}_2)/2$ and derive heading and speed references:
\begin{align}
	\theta_{\text{ref}} &= \operatorname{atan2}(a_y,a_x),\\
	v_{\text{ref}} &= \gamma \,\big\Vert \hat{\mathbf{w}}_1-\hat{\mathbf{w}}_2 \big\Vert_2,
\end{align}
with a scale $\gamma>0$. Let $v$ be the measured linear speed. Lateral and longitudinal PID outputs are
\begin{align}
	u_{\text{lat}} &= \mathrm{PID}_{\text{lat}}(\,\theta_{\text{ref}}-\theta\,),\\
	u_{\text{lon}} &= \mathrm{PID}_{\text{lon}}(\,v_{\text{ref}}-v\,).
\end{align}

We convert them into \emph{position–orientation} control $(x,y,\theta)$, e.g., by a steering–throttle mapping:
\begin{equation}
	u^{\text{PID}} = (x_{\text{PID}},y_{\text{PID}},\theta_{\text{PID}})=\Psi(u_{\text{lat}},u_{\text{lon}}).
\end{equation}

\textbf{Command–specific MLP controllers.}
In parallel, \emph{command–specific} MLPs infer control directly from $\mathbf{h}_t$ (not from waypoints).
We first infer a discrete command $C_t\in\{\,\text{straight},\text{left},\text{right}\,\}$ from the upcoming two route points (in local coordinates $(R^x_{p1},R^x_{p2})$), using simple thresholds:
\begin{equation}
	C_t=
	\begin{cases}
		\text{left},  & R^x_{p1}\le -\tau_1 \ \lor\ R^x_{p2}\le -\tau_2,\\[2pt]
		\text{right}, & R^x_{p1}\ge  \tau_1 \ \lor\ R^x_{p2}\ge  \tau_2,\\[2pt]
		\text{straight}, & \text{otherwise}.
	\end{cases}
\end{equation}

\noindent with an MLP head specific to $C_t$ predicts
\begin{equation}
	u^{\text{MLP}} = (x_{\text{MLP}},y_{\text{MLP}},\theta_{\text{MLP}})=\mathrm{MLP}_{C_t}(\mathbf{h}_t).
\end{equation}

\textbf{Control blending policy.}
The final control blends both controllers with confidence gating (threshold $\epsilon>0$) and weights $\beta_{ij}\in[0,1]$:
\begin{align}
	\label{eq:blend}
	\text{if}\ \|u^{\text{MLP}}\|\ge\epsilon\ \wedge\ \|u^{\text{PID}}\|\ge\epsilon:\
	u_t &= \begin{bmatrix}
		\beta_{00} & \beta_{10}\\
		\beta_{01} & \beta_{11}
	\end{bmatrix}
	\!\begin{bmatrix}
		u^{\text{MLP}}\\
		u^{\text{PID}}
	\end{bmatrix};\\
	\text{else if}\ \|u^{\text{MLP}}\|\ge\epsilon:&\ u_t=u^{\text{MLP}};\\
	\text{else if}\ \|u^{\text{PID}}\|\ge\epsilon:&\ u_t=u^{\text{PID}};\\
	\text{else}:&\ u_t=\mathbf{0},
\end{align}

\noindent where $\beta_{ij}$ are task loss weight which are tuned adaptively during training process (see Subsection \ref{subsec:training}). Algorithm \ref{alg:policy} summarizes how the latent spaces from the perception encoder are translated into low-level control actions. \revz{Note that our 'end-to-end' scope refers to the learned navigation policy. The robot's built-in low-level controller handles the locomotion dynamics and inverse kinematics required to execute these commands ($u_t: u^{\text{MLP}}$ and/or $u^{\text{PID}}$) via leg movements.}

\begin{algorithm}[t]
	\caption{Control Policy}
	\label{alg:policy}
	\DontPrintSemicolon
	\SetAlgoLined
	\small
	\KwIn{Perception latent $\mathbf{z}_t$,\\
		Previous GRU state $\mathbf{h}_{t-1}$,\\ 
		Route points $(R^x_{p1},R^x_{p2})$}
	\KwOut{Control $u_t=(x,y,\theta)$}
	--------------------------------------------------------------------\\
	Compute $\mathbf{h}_t$ with GRU from $\mathbf{z}_t$\;
	Roll out waypoints $\{\hat{\mathbf{w}}_\ell\}_{\ell=1}^N$\;
	$u^{\text{PID}} \leftarrow \Psi(\mathrm{PID}_{\text{lat/lon}}(\hat{\mathbf{w}}_{1:2}))$\;
	Infer command $C_t$ from $(R^x_{p1},R^x_{p2})$\;
	$u^{\text{MLP}} \leftarrow \mathrm{MLP}_{C_t}(\mathbf{h}_t)$\;
	Blend $u^{\text{PID}}$ and $u^{\text{MLP}}$ using Eq.~\eqref{eq:blend}\;
	\Return $u_t$\;
\end{algorithm}

\subsubsection{Global-to-Local Transformation}
The model ingests, besides RGB-D and GNSS, a \emph{sequence of route points} that prescribes the path from start to goal. At time $t$, we select the next two route points $(\phi^{(1)},\lambda^{(1)})$, $(\phi^{(2)},\lambda^{(2)})$ (global lat--lon) and transform them to local Cartesian coordinates $(R^x_{p1},R^y_{p1})$, $(R^x_{p2},R^y_{p2})$ in the BEV frame where the robot sits at $(0,0)$ (bottom center).

First compute the bearing $\beta$ from two consecutive robot GNSS positions $(\phi_1,\lambda_1)$ and $(\phi_2,\lambda_2)$:
\begin{equation}
	\begin{aligned}
		\beta = \operatorname{atan2}\big(&\,\sin\Delta\lambda \,\cos\phi_{2},\\
		&\,\cos\phi_{1}\sin\phi_{2}-\sin\phi_{1}\cos\phi_{2}\cos\Delta\lambda\big),
	\end{aligned}
	\label{eq:bearing}
\end{equation}

\begin{figure}[!t]
	\begin{center}
		\includegraphics[width=\linewidth]{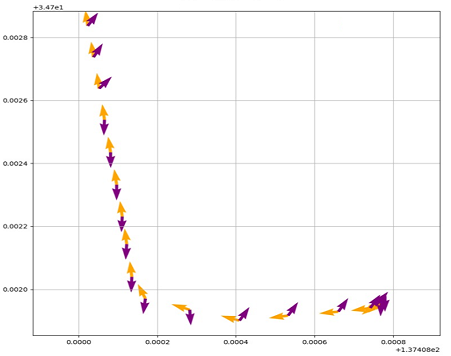}
		
	\end{center}
	%\vspace{-4mm}
	\caption{GNSS-based bearing estimation (orange) vs 9-axis IMU with EKF-based bearing estimation (purple).}
	\label{fig:gnssvsimu}
	%\vspace{-3mm}
\end{figure}

%\rev{To validate our design choice, Fig.~\ref{fig:gnssvsimu} compares the GNSS-based bearing (orange) against the heading derived from the external 9-axis Witmotion IMU (purple). The IMU orientation exhibits characteristic drift due to magnetic interference (hard/soft iron effects) common in urban environments. In contrast, the GNSS-based estimation, while noisier at low speeds, maintains global consistency without drift. Consequently, we rely on the GNSS bearing and utilize the GRU to smooth the high-frequency noise.}
\noindent with $\Delta\lambda=\lambda_2-\lambda_1$. \revx{While fusing IMU and vision is standard for local odometry (VIO), relying on magnetometers for \textit{global} absolute heading is prone to significant errors in urban settings due to hard- and soft-iron magnetic interference from buildings and infrastructure \cite{magrobot}. To validate our design choice, Fig.~\ref{fig:gnssvsimu} compares the GNSS-based bearing estimation against the heading derived from the external 9-axis Witmotion IMU. The magnetometer-based heading (purple) drifts significantly over time. To ensure global consistency for route-following, we explicitly discard the magnetometer and derive absolute bearing from the differential GNSS signal (orange). While the raw GNSS bearing is noisier at low speeds, it is immune to magnetic drift. Our GRU-based control policy effectively smooths this high-frequency noise, combining the drift-free nature of GNSS with the temporal stability of the recurrent network.} However, performance degraded near tall buildings due to partial satellite blockage, consistent with our qualitative observations. For a nearby route point $(\phi_r,\lambda_r)$ and current robot fix $(\phi_c,\lambda_c)$, an equirectangular approximation yields the local offsets (meters)
\begin{align}
	\Delta x' &= C_e\cos\phi_c\,(\lambda_r-\lambda_c),\qquad
	\Delta y' = C_m(\phi_r-\phi_c),\\
	\begin{bmatrix}R^x_p\\ R^y_p\end{bmatrix}
	&=
	\begin{bmatrix}\cos\beta&-\sin\beta\\ \sin\beta&\cos\beta\end{bmatrix}
	\begin{bmatrix}\Delta x'\\ \Delta y'\end{bmatrix},
\end{align}

\begin{table}[t] %[hb]
	\caption{Statistics of the Seq-DeepIPC Dataset}
	%\vspace{-5mm}
	\begin{center}
		\resizebox{\linewidth}{!}{%
			%\begin{tabular*}{\textwidth}{@{\extracolsep{\stretch{1}}}*{7}{r}@{}}
			\begin{tabular}{p{0.275\linewidth}p{0.65\linewidth}}
				% \begin{tabular}{cc}
					\toprule
					% \hdashline\noalign{\vskip 0.75ex}
					
					Conditions & Sunny and Cloudy\\
					\midrule
					Total routes & 16 (train), 5 (validation), 5 (test)\\
					\midrule
					%Train frames & 4831 (noon), 5320 (evening), 10151 (total)\\
					%Val frames & 4863 (noon), 4816 (evening), 9679 (total)\\
					%Test frames & 9510 (noon), 9465 (evening), 18975 (total)\\
					$\mathcal{N}$ samples* & 5164 (train), 1799 (val), 1681 (test)\\
					\midrule
					Devices & \rev{Unitree Go2 Legged Robot}\\
					& \rev{Stereolabs Zed 2i RGBD camera}\\
					& \rev{U-blox Zed-F9P RTK-GNSS receiver}\\
					& \rev{9-axis Witmotion HWT905 IMU (only for baseline comparison, see Fig. \ref{fig:gnssvsimu})}\\
					
					% \hdashline\noalign{\vskip 0.75ex}
					\midrule
					Object classes & Following Cityscapes classes ~\cite{Cordts2016Cityscapes}\\
					% \hdashline\noalign{\vskip 0.75ex}
					
					\bottomrule                             
				\end{tabular}
			}
		\end{center}
		\label{tab:dataset}
		%\vspace{-2mm}
		\begin{tablenotes}\small
			\item *$\mathcal{N}$ samples is the number of observation sets. Each set consists of an RGBD image, GNSS location, and control signals.
		\end{tablenotes}
		%\vspace{-1mm}
	\end{table}

\noindent where \(C_e\) and \(C_m\) are the local radii of curvature of the Earth ellipsoid at the current latitude \(\phi_c\), given by  
\[
C_m = \frac{a(1-e^2)}{(1 - e^2 \sin^2 \phi_c)^{3/2}}, 
\qquad
C_e = \frac{a}{\sqrt{1 - e^2 \sin^2 \phi_c}},
\]
with \(a\) the semi-major axis and \(e\) the eccentricity of the reference ellipsoid as in the WGS84 standard parameters, with $a = 6378137 \,\text{m}$ and $\qquad e^2 = 0.00669437999014$. This formulation accounts for the Earth’s ellipsoidal shape, improving accuracy over the spherical approximation. It ensures precise projection of the route points into the local frame aligned with the BEV map. This alignment enables: (i) command inference from \((R^x_{p1}, R^x_{p2})\) (left/straight/right), and (ii) conditioning the GRU on the path heading relative to the robot.

 %In practice, we adopt the WGS84 standard parameters:  
%\[
%a = 6378137 \,\text{m}, \qquad e^2 = 0.00669437999014.
%\]

\subsection{Dataset}

	\begin{figure}[!t]
		\begin{center}
			\includegraphics[width=\linewidth]{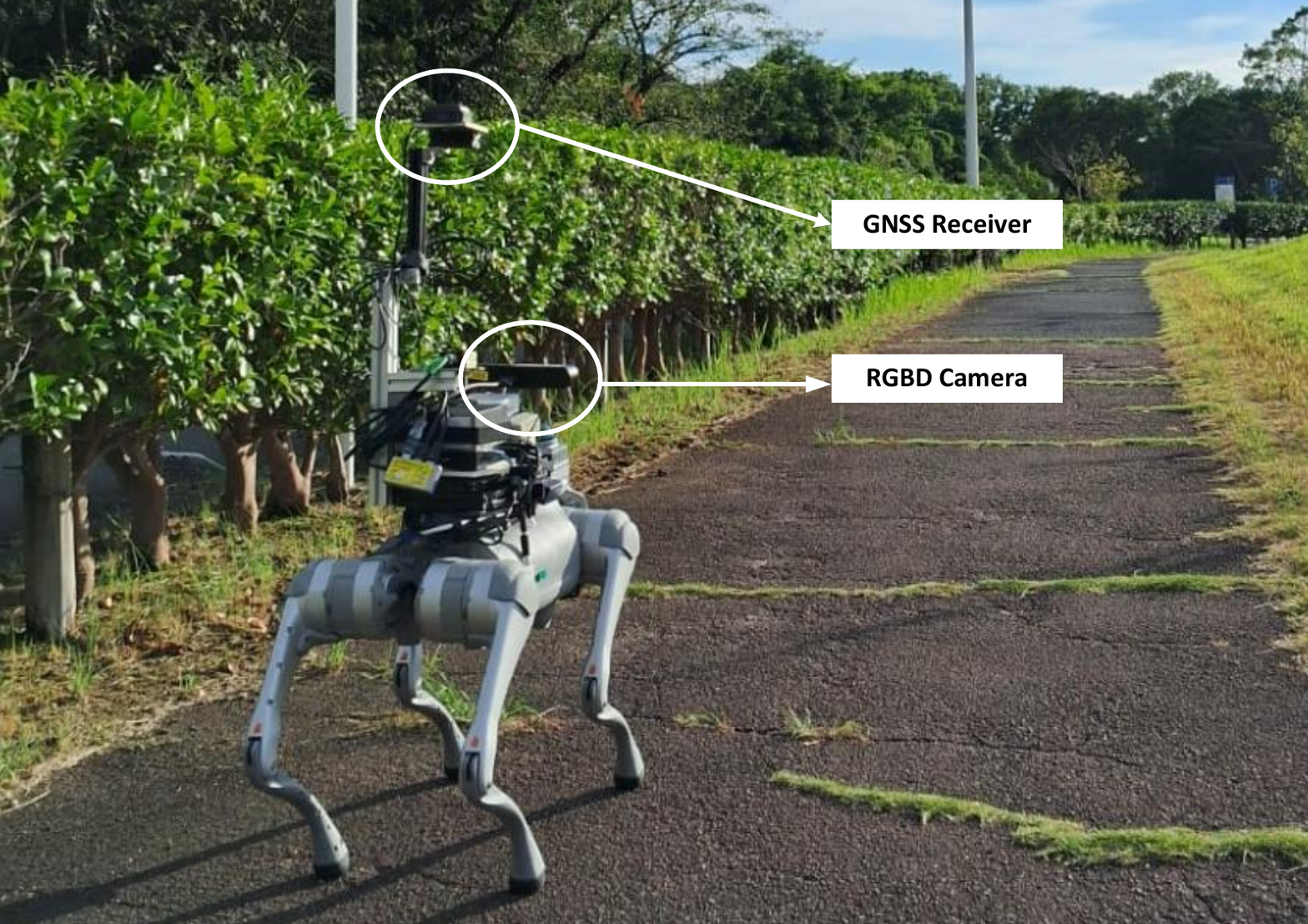}
			
		\end{center}
		%\vspace{-4mm}
		\caption{Sensor placement on the legged robot (Unitree Go2).} %  
		\label{fig:robotsensor}
		%\vspace{-4mm}
	\end{figure}

	\begin{figure}[!t]
		\begin{center}
			\includegraphics[width=\linewidth]{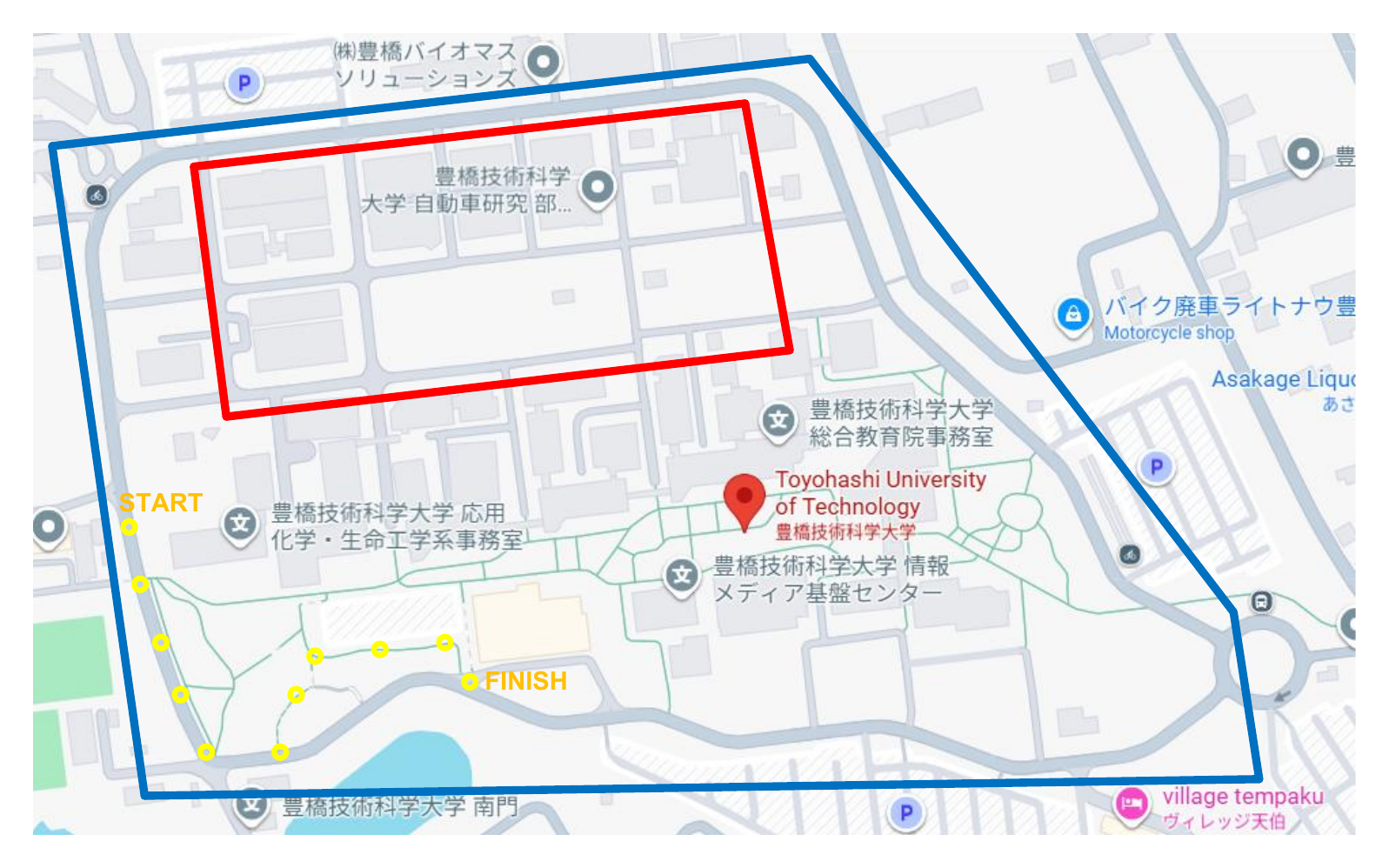}
			
		\end{center}
		%\vspace{-4mm}
		\caption{The experiment area. Red: Old DeepIPC dataset coverage area. Blue: extended coverage area used for all models in this experiment. Yellow hollow circles represent a route that consists of start, finish, and a set of route points. (\href{https://goo.gl/maps/9rXobdhP3VYdjXn48}{https://goo.gl/maps/9rXobdhP3VYdjXn48})}
		\label{fig:route}
		%\vspace{-3mm}
	\end{figure}

The dataset for Seq-DeepIPC was collected at Toyohashi University of Technology, Japan, over an extended campus route that includes both structured road surfaces and unstructured grassy areas. Compared to the original DeepIPC dataset (shorter loop, mainly road surfaces), the new dataset covers a larger perimeter and more diverse conditions, which better reflects the challenges of legged robot navigation. The summary of the dataset and sensor information can be seen on Table \ref{tab:dataset} while the sensor placement on the legged robot can be seen on Fig. \ref{fig:robotsensor}. A schematic of the experiment area is shown in Fig.~\ref{fig:route}. The area inside the red lines is the coverage area of the old DeepIPC dataset. Meanwhile, the area inside the blue lines is the extended coverage area.

In total, the dataset consists of 26 distinct trajectories, partitioned into 16 routes for training, 5 routes for validation, and 5 routes for testing. Each trajectory contains synchronized multimodal streams:
\begin{itemize}
	\item RGB images $I^R_t$ and depth maps $I^D_t$ captured at 30~FPS,
	\item GNSS measurements $(\phi_t,\lambda_t)$ at 1~Hz,
	\item route point sequences $\mathcal{P}=\{(\phi^{(m)},\lambda^{(m)})\}$ that define the global path,
	\item expert control commands $u_t=(x,y,\theta)$ recorded from teleoperation of the legged robot, and
	\item \rev{estimated forward velocity $v_t$ obtained directly from the GNSS receiver. Unlike velocity calculated via position differencing, the receiver estimates velocity using Doppler shift measurements, providing higher accuracy for the control policy input.}
\end{itemize}

%\smallskip
\textbf{Ground-truth construction.}
The ground-truth waypoints are extracted from the robot’s actual traversal trajectory during data collection. This trajectory is estimated using a visual–inertial odometry (VIO) algorithm integrated into the robot’s on-board edge device, providing locally consistent 6-DoF motion estimates. To be noted, we only consider 5 future waypoints to be predicted; they only have a gap of around 5 meters (between the robot's current location to the fifth waypoint). Thus, VIO was preferred for trajectory ground-truth construction because it provides denser and locally more consistent data than GNSS, which can suffer from signal degradation and update delays. While GNSS is used for heading estimation during deployment, VIO yields smoother reference trajectories for supervised waypoint prediction.

The ground-truth position–orientation controls $(x,y,\theta)$ correspond to the remote control states issued during teleoperation, ensuring that the model learns from human-expert steering actions.  
Depth supervision uses the synchronized RGB-D measurements $I^D_t$, whereas semantic segmentation ground truth is obtained using the pretrained SegFormer~\cite{xie2021segformer} model. SegFormer, trained on the Cityscapes dataset~\cite{Cordts2016Cityscapes}, provides strong generalization to outdoor scenes and acts as a ``teacher'' model in a knowledge-distillation manner, where Seq-DeepIPC serves as the ``student''.

\smallskip
To ensure temporal consistency, frames are grouped into short observation windows of length $K\in\{1,2,3\}$, producing training samples of the form
\begin{equation}
	(O_t, P_t, g_t; S_t, D_t, W_t, u_t),
	\label{eq:dataset_sample}
\end{equation}
where $O_t$ is the sequence of RGB-D images, $P_t$ the sequence of route points that will be transformed into local coordinates, $g_t$ is the sequence of GNSS measurements, and $\{S_t,D_t,W_t,u_t\}$ are the ground-truth labels supervising the multi-task outputs.

\subsection{Training Configuration}
\label{subsec:training}

We train Seq-DeepIPC using supervised imitation learning on a PC equipped with an RTX~4090 GPU. The model that achieves the lowest validation $\mathcal{L}_{\text{total}}$ is selected for testing. We adopt the AdamW optimizer \cite{adamw} with $(\beta_1,\beta_2)=(0.9,0.999)$, an initial learning rate of $\eta_0=10^{-4}$, and weight decay of $10^{-4}$. The batch size is set to~5. Training continues until early stopping is triggered when the validation $\mathcal{L}_{\text{total}}$ shows no improvement for~30 consecutive epochs. The learning rate follows a step decay schedule: if no improvement occurs for~5 epochs, the learning rate is halved, following $\eta\leftarrow\max(0.5\eta,\,\eta_{\min})$ with $\eta_{\min}=10^{-6}$. The sequence length for temporal inputs is $K\in\{1,2,3\}$. \rev{We selected $K=3$ as the optimal trade-off based on empirical testing. Shorter sequences ($K<3$) failed to adequately smooth the camera shake caused by the robot's stepping gait.}

\textbf{Overall training objective.}
The total loss combines perception, waypoint, and control objectives. The overall loss is a weighted sum of task-specific losses:
\begin{equation}
	\mathcal{L}_{\text{total}} =
	\alpha_{percep}\,\mathcal{L}_{percep} +
	\alpha_{wp}\,\mathcal{L}_{wp} +
	\alpha_{ctrl}\,\mathcal{L}_{ctrl},
\end{equation}
where the task weights $\alpha_{\{\cdot\}}$ are automatically tuned using Modified Gradient Normalization (MGN) \cite{mgn} as in DeepIPC. This ensures equitable gradient magnitudes across tasks, preventing dominance by any single loss and promoting stable multi-task learning.

\begin{table*}[t] %[hb]
\caption{Model Specification}
%\vspace{-5mm}
\begin{center}
	\resizebox{\textwidth}{!}{%
		%\begin{tabular*}{\textwidth}{@{\extracolsep{\stretch{1}}}*{7}{r}@{}}
		% \begin{tabular}{|p{0.2\linewidth}|p{0.2\linewidth}|p{0.1\linewidth}|p{0.4\linewidth}|}
			\begin{tabular}{ccccc}
				%\toprule
				%& $z_{6}$ & $z_{8}$ & $z_{9}$ & $z_{11}$ & $z_{13}$ & $z_{14}$ \\
				%\midrule
				% \hline
				\toprule
				% {\bf Model} & {\bf Total Parameters$\downarrow$} & {\bf GPU Load (MB)$\downarrow$} & {\bf Input/Sensor}\\
				Model & Total Parameters$\downarrow$ & Model Size $\downarrow$ & Input/Sensor & Output\\
				\toprule
				% \hline
				Huang \cite{huang_model} & 74953360 & 300.2 MB & RGBD, High-level commands & Segmentation, Low-level Controls\\
				% \hline
				AIM-MT \cite{aim_mt} & 27967063 & 112.1 MB & RGB, GNSS & Segmentation, Depth, Waypoints\\
				% \hline
				% \hdashline\noalign{\vskip 0.75ex}
				DeepIPC \cite{natan2024deepipc} & 20953266 & 85.0 MB & RGBD, GNSS & Segmentation, BEV Semantic, Waypoints, Low-level Controls\\
				% \hline
				% \hdashline\noalign{\vskip 0.75ex}
				Seq-DeepIPC & 12291227 & 47.5 MB & RGBD, GNSS & Segmentation, Depth, BEV Semantic, Waypoints, Low-level Controls\\
				% \hline
				\bottomrule                             
			\end{tabular}
		}
	\end{center}
	%\vspace{-1mm}
	\label{tab:model_compare}
	\begin{tablenotes}\small
		%\item *These outputs are not supervised with a certain loss function during training process.
		%
		\item We implement Huang's model \cite{huang_model} based on their paper. Meanwhile, AIM-MT \cite{aim_mt} and DeepIPC \cite{natan2024deepipc} are implemented based on author's original repository that can be accessed at \href{https://github.com/autonomousvision/neat}{https://github.com/autonomousvision/neat} and \href{https://github.com/oskarnatan/DeepIPC}{https://github.com/oskarnatan/DeepIPC} with a small modification for controlling a legged robot. All models are deployed on a Jetson AGX Orin.
	\end{tablenotes}
	%\vspace{-2mm}
\end{table*}

\paragraph*{Segmentation loss}
Semantic segmentation is trained using an additive combination of Binary Cross-Entropy (BCE) and Dice loss:
\begin{equation}
	\mathcal{L}_{\text{seg}} = \text{BCE}(\hat{S}, S) + \text{Dice}(\hat{S}, S),
	\label{eq:seg_loss}
\end{equation}
where
\begin{equation}
	\text{BCE}(\hat{S}, S) = -\frac{1}{N}\sum_{i=1}^{N} \big(S_i \log \hat{S}_i + (1 - S_i)\log(1 - \hat{S}_i)\big),
\end{equation}
\begin{equation}
	\text{Dice}(\hat{S}, S) = 1 - \frac{2\sum_i \hat{S}_i S_i + \epsilon}{\sum_i \hat{S}_i + \sum_i S_i + \epsilon}.
\end{equation}

The BCE term encourages pixel-wise classification accuracy and is particularly effective for well-balanced regions, while the Dice loss improves overlap-based similarity, mitigating class imbalance and emphasizing small or thin structures in the scene. The combination allows the network to optimize both global segmentation consistency and region-wise precision.

\paragraph*{Depth loss}
Depth estimation is optimized using a combination of $\ell_1$ and $\ell_2$ norms:
\begin{equation}
	\mathcal{L}_{\text{depth}} = \|\hat{D} - D\|_1 + \|\hat{D} - D\|_2^2.
\end{equation}

The $\ell_1$ component enforces robustness against outliers and preserves local discontinuities (e.g., edges and object boundaries), whereas the $\ell_2$ term penalizes large residuals more strongly, encouraging overall smoothness and stable convergence. Their combination balances detail preservation and global consistency, which is critical for learning depth maps that support reliable BEV projection.

\paragraph*{Perception loss}
The perception branch jointly optimizes both segmentation and depth estimation objectives:
\begin{equation}
	\mathcal{L}_{\text{percep}} = \mathcal{L}_{\text{seg}} + \mathcal{L}_{\text{depth}}.
\end{equation}

This multi-task formulation enables the encoder to learn shared spatial representations that improve downstream control and planning accuracy.

\paragraph*{Waypoint and control losses}
Both waypoint regression and control prediction are formulated as continuous regression tasks, optimized using a combination of $\ell_1$ and $\ell_2$ losses:
\begin{align}
	\mathcal{L}_{\text{wp}} &= \|\hat{W} - W\|_1 + \|\hat{W} - W\|_2^2, \label{eq:wp_loss}\\
	\mathcal{L}_{\text{ctrl}} &= \|\hat{u} - u\|_1 + \|\hat{u} - u\|_2^2. \label{eq:ctrl_loss}
\end{align}

The $\ell_1$ component provides robustness to noisy teleoperation labels and occasional trajectory deviations, ensuring stable learning from imperfect demonstrations, while the $\ell_2$ term enforces smooth convergence and penalizes large prediction errors in position and orientation. This dual objective allows the controller to maintain both trajectory precision and smooth actuation responses which are critical properties for legged locomotion in unstructured terrain.

\subsection{Evaluation Settings}
%and efficiency
We adopt both offline and online evaluations. Offline tests provide quantitative accuracy while online tests verify qualitative behavior on the real legged robot.

\textbf{Offline tests.} \revx{To rigorously validate our architectural decisions, we benchmark Seq-DeepIPC against three baselines strategically selected to isolate specific design components. DeepIPC \cite{natan2024deepipc} serves as a single-frame ablation, isolating the performance gain specifically attributable to temporal sequence modeling. Huang et al. \cite{huang_model} represents sensor fusion without recurrent memory, allowing us to quantify the benefit of the GRU for stabilizing legged robot jitter. Finally, AIM-MT \cite{aim_mt} represents multi-task learning without explicit BEV projection, isolating the benefit of the geometry-informed spatial transformation. This comparative set effectively disentangles the contributions of sequentiality, recurrence, and spatial representation. We also create variants of theirs that consume the same inputs as ours for deeper analysis in the comparative study.}

The model specification details can be seen on Table \ref{tab:model_compare}. It shows that our proposed model, Seq-DeepIPC is lighter than other models in terms of number of trainable parameters and model size. Since not all models predict the same outputs (e.g., DeepIPC does not predict depth), comparisons are reported task-wise rather than in a single aggregate metric. We also ablate the sequence length $K\in\{1,2,3\}$. All models are evaluated on the $5$ held-out test routes using metrics: segmentation IoU, depth MAE, waypoint MAE, and the high-level control commands MAE.

\begin{table*}[t]
	\centering
	\caption{Quantitative comparison of Seq-DeepIPC and baselines. Results are reported as mean $\pm$ standard deviation across three runs. Best values in each column are highlighted in bold.}
	\begin{tabular}{lccccc} %c
		\hline
		Model & Seq Input & Seg. IoU $\uparrow$ & Depth MAE $\downarrow$ & WP MAE $\downarrow$ & Ctrl MAE $\downarrow$ \\% & FPS $\uparrow$ \\
		\hline
		Huang \cite{huang_model} & 1 & 0.814 $\pm$ 0.003 & -- & -- & 0.061 $\pm$ 0.001 \\%& \textbf{129.5 $\pm$ 15.1} \\
		& 2 & 0.807 $\pm$ 0.005 & -- & -- & 0.060 $\pm$ 0.001 \\%& 64.2 $\pm$ 5.3 \\
		& 3 & 0.809 $\pm$ 0.003 & -- & -- & 0.062 $\pm$ 0.004 \\%& 44.7 $\pm$ 4.9 \\
		\hline
		AIM-MT \cite{aim_mt} & 1 & 0.847 $\pm$ 0.004 & 0.091 $\pm$ 0.002 & 0.710 $\pm$ 0.025 & -- \\%& 112.2 $\pm$ 14.6 \\
		& 2 & 0.836 $\pm$ 0.018 & 0.090 $\pm$ 0.001 & 0.713 $\pm$ 0.015 & -- \\%& 59.5 $\pm$ 6.0 \\
		& 3 & 0.842 $\pm$ 0.003 & 0.091 $\pm$ 0.003 & \textbf{0.714 $\pm$ 0.026} & -- \\%& 41.1 $\pm$ 4.6 \\
		\hline
		DeepIPC \cite{natan2024deepipc} & 1 & 0.839 $\pm$ 0.002 & -- & 0.770 $\pm$ 0.017 & 0.055 $\pm$ 0.003 \\%& 37.7 $\pm$ 0.8 \\
		& 2 & 0.839 $\pm$ 0.004 & -- & 0.739 $\pm$ 0.033 & 0.057 $\pm$ 0.007 \\%& 19.6 $\pm$ 0.8 \\
		& 3 & \textbf{0.847 $\pm$ 0.001} & -- & 0.729 $\pm$ 0.019 & 0.047 $\pm$ 0.001 \\%& 13.3 $\pm$ 0.3 \\
		\hline
		Seq-DeepIPC & 1 & 0.839 $\pm$ 0.003 & 0.088 $\pm$ 0.005 & 0.774 $\pm$ 0.039 & 0.063 $\pm$ 0.009 \\%& 44.5 $\pm$ 2.1 \\
		& 2 & 0.837 $\pm$ 0.006 & 0.087 $\pm$ 0.003 & 0.730 $\pm$ 0.019 & 0.058 $\pm$ 0.002 \\%& 23.7 $\pm$ 0.7 \\
		& 3 & 0.844 $\pm$ 0.002 & \textbf{0.084 $\pm$ 0.001} & 0.725 $\pm$ 0.010 & \textbf{0.046 $\pm$ 0.001} \\%& 16.0 $\pm$ 0.6 \\
		\hline
	\end{tabular}
	\label{tab:quant_results}
\end{table*}

\paragraph*{Segmentation IoU}
Let $\mathcal{C}$ be the set of semantic classes and $\Omega$ the pixel domain.
For class $c\!\in\!\mathcal{C}$, with ground truth $S^{(c)}\!\in\!\{0,1\}^{\Omega}$ and prediction $\hat{S}^{(c)}\!\in\![0,1]^{\Omega}$, % thresholded to $\{0,1\}$,
\begin{equation}
	\mathrm{IoU}_c \;=\; \frac{\left\lvert\, \hat{S}^{(c)} \cap S^{(c)} \,\right\rvert}{\left\lvert\, \hat{S}^{(c)} \cup S^{(c)} \,\right\rvert},
\end{equation}
the IoU is
\begin{equation}
	\mathrm{IoU} \;=\; \frac{1}{|\mathcal{C}|}\sum_{c\in\mathcal{C}}\mathrm{IoU}_c.
\end{equation}

\paragraph*{Depth MAE}
For per-pixel depth $D,\hat{D}\in\mathbb{R}^{\Omega}$ (meters),
\begin{equation}
	\mathrm{MAE}_{\text{depth}} \;=\; \frac{1}{|\Omega|} \sum_{p\in\Omega} \left|\hat{D}(p)-D(p)\right|.
\end{equation}

\paragraph*{Waypoint MAE}
Given $N$ local waypoints, $W=\{\mathbf{w}_\ell\}_{\ell=1}^{N}$ and $\hat{W}=\{\hat{\mathbf{w}}_\ell\}_{\ell=1}^{N}$,
\begin{equation}
	\mathrm{MAE}_{\text{wp}} \;=\; \frac{1}{N}\sum_{\ell=1}^{N} \left\| \hat{\mathbf{w}}_\ell - \mathbf{w}_\ell \right\|_1 .
\end{equation}

\paragraph*{Control MAE}
For robot control $u=(x,y,\theta)\in\mathbb{R}^{3}$,
\begin{equation}
	\mathrm{MAE}_{\text{ctrl}} \;=\; \frac{1}{3}\Big( \left| \hat{x}-x \right| + \left| \hat{y}-y \right| + \left| \hat{\theta}-\theta \right| \Big).
\end{equation}

\textbf{Online tests (qualitative).} 
We deploy Seq\mbox{-}DeepIPC on a Unitree legged robot across predefined campus routes that include both asphalt roads and grassy terrain. Unlike the wheeled platform used in DeepIPC, the legged robot can safely traverse uneven and semi-structured surfaces without manual intervention. We further analyze challenging conditions such as partial GNSS occlusion near tall buildings. In these regions, satellite signal degradation perturbs the bearing estimation and consequently the global-to-local coordinate transformation, resulting in gradual navigation drift. 
Qualitative evaluation focuses on visualizing the full perception-to-control pipeline, including RGB inputs, predicted segmentation and depth maps, BEV projections, planned waypoints, and $(x, y, \theta)$ control traces overlaid with the ground-truth GNSS trajectory. These visualizations reveal that Seq-DeepIPC produces temporally consistent segmentations and smooth waypoint transitions, leading to stable locomotion even under texture variation and illumination changes. %

\begin{figure*}[t]
	\centering
	\subfloat[Open area, road traversal (successful)]{
		\includegraphics[width=\textwidth]{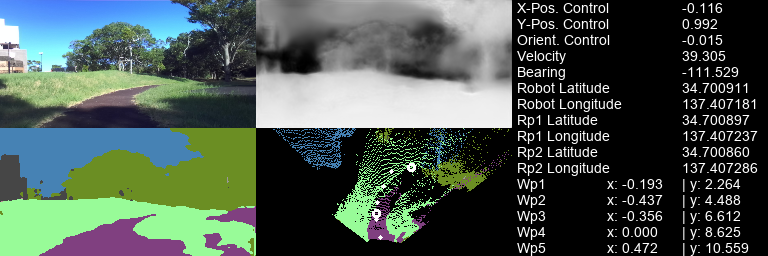} 
	}
	
	\subfloat[Open area, grass (stairs) traversal (successful)]{
		\includegraphics[width=\textwidth]{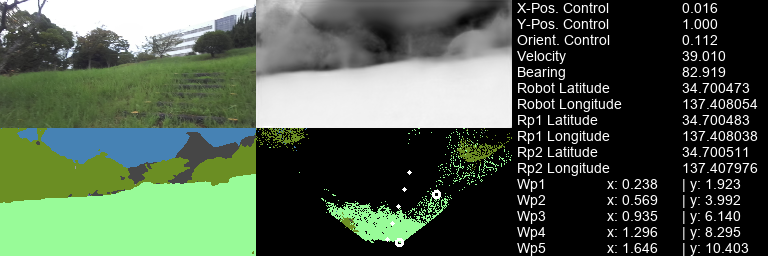} 
	}
	
	\subfloat[Failure near tall building]{
		\includegraphics[width=\textwidth]{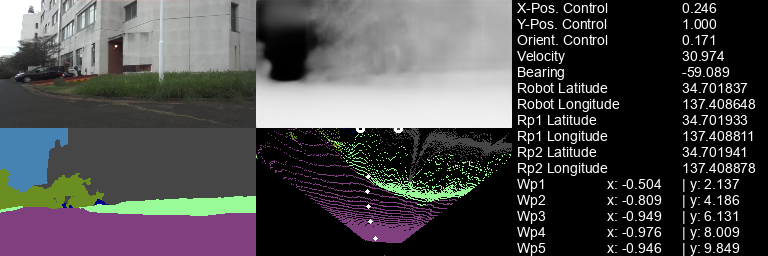}
	}
	
	\caption{Qualitative results of Seq-DeepIPC deployment on a Unitree legged robot. Each row corresponds to an observation set, showing representative outputs. Negative x-pos and orient. controls mean to the left, while positive means to the right. (a) Successful road traversal, (b) successful grass (stairs) traversal, (c) failure case near tall buildings, as the model fails to predict waypoints correctly due to misplaced route points.}
	\label{fig:qual_results}
	%\vspace{-2.5pt}
\end{figure*}

\section{Results and Discussion}

In this research, we comprehensively evaluate Seq-DeepIPC and three other baselines: DeepIPC, AIM-MT, and Huang through both offline quantitative analysis and online real-world deployment. Offline evaluation assesses perception and control using four metrics: IoU for segmentation and MAE for depth, waypoint, and control outputs $(x, y, \theta)$. Each model is trained and tested three times with randomized seeds, and mean $\pm$ std values are reported for statistical reliability. The consolidated quantitative results are summarized in Table~\ref{tab:quant_results}. Online evaluation further deploys the best-performing Seq-DeepIPC on a Unitree Go2 legged robot operating on both asphalt and grass terrains, using live RGB-D and GNSS inputs processed on a Jetson AGX Orin. Qualitative outcomes, shown in Fig.~\ref{fig:qual_results}, confirm that Seq-DeepIPC can maintain smooth navigation and robust perception across heterogeneous real-world environments. Qualitative results are discussed in more detail in Sec.~\ref{sec:qualitative}.

\subsection{Ablation Study and Sequential Effects}

\textbf{Impact of sequential inputs.}
The number of sequential frames $K$ markedly affects models with temporal encoders (DeepIPC, Seq-DeepIPC) but not those lacking explicit recurrence (AIM-MT, Huang). As shown in Table \ref{tab:quant_results}, increasing $K$ from 1 to 3 yields monotonic gains in IoU, depth accuracy, waypoint precision, and control smoothness for both DeepIPC variants. This trend empirically confirms that the GRU-based latent integration captures temporal continuity across frames, suppressing per-frame perception noise (e.g., short occlusions). In contrast, Huang and AIM-MT process stacked frames independently; thus, temporal redundancy provides little additional information and may even cause feature over-smoothing. The results highlight that merely feeding sequential frames is insufficient, so that architectural mechanisms must translate temporal correlation into a stable latent state.

\textbf{Temporal dynamics and control stability.}
Seq-DeepIPC’s GRU stabilizes the latent representations by integrating temporal dependencies across consecutive RGB-D frames. As a result, the latent features fed to the control policy are temporally consistent, reducing erratic command outputs. In practice, this temporal coherence suppresses overreactive PID corrections and leads to more stable actuation, particularly in mixed-terrain navigation where both visual and depth signals exhibit higher variance. Quantitatively, the reduction in control MAE from $K{=}1$ to $K{=}3$ is the most significant among all ablation settings. This suggests that Seq-DeepIPC does not merely memorize local transitions but effectively models short-term dynamics for perception stability and motion smoothness.

\subsection{Effect of Multi-Task Perception and BEV Fusion}

Seq-DeepIPC's dual supervision (multi-task semantic segmentation and depth estimation) encourages the encoder to preserve geometry-aware features useful for both perception and planning. The auxiliary loss leads to latent spaces with improved spatial consistency. \rev{Quantitatively, this design choice is supported by the comparison in Table \ref{tab:quant_results}. First, regarding \textbf{Multi-Task Learning}: although Seq-DeepIPC uses a significantly lighter encoder (EfficientNet-B0) compared to the original DeepIPC (EfficientNet-B3) and Huang (ResNet-50), it achieves comparable or superior performance, particularly at $K=3$. This indicates that the auxiliary depth supervision effectively enriches the latent features, compensating for the reduced model capacity and acting as a necessary structural prior for navigation. Second, regarding \textbf{BEV Fusion}: Seq-DeepIPC significantly outperforms the baseline by Huang \cite{huang_model}, which relies only on perspective-view fusion. This performance gap confirms that BEV map provides a more geometrically consistent state space for the controller, simplifying the mapping from raw sensor data to high-level motion commands.}

Furthermore, by jointly learning segmentation and depth estimation, the model disentangles texture and structural cues, enabling better obstacle boundary awareness and traversability reasoning. The joint segmentation–depth learning also reduces over-fitting: gradients from the two tasks balance semantic and geometric fidelity. This complementary effect explains why Seq-DeepIPC consistently outperforms DeepIPC, particularly when sequential information is abundant ($K{=}3$). Depth estimation forces the encoder to exploit subtle photometric cues, while segmentation enforces semantic boundary awareness. Together, they produce robust features that generalize across terrain textures which is essential for accurate navigation on uneven surfaces.

\subsection{Task-Wise Comparative Analysis.}
\textbf{Perception accuracy} AIM-MT, DeepIPC, and Seq-DeepIPC achieve significantly higher IoU than Huang's model, confirming the benefit of multi-task and feature-fusion in their network architecture. Despite employing a lighter encoder, Seq-DeepIPC maintains near state-of-the-art IoU, validating that its temporal and depth cues effectively compensate for reduced model size. The sequential variant produces more stable segmentation boundaries and fewer transient misclassifications across frames. This stability stems from GRU-based temporal smoothing, which suppresses frame-to-frame fluctuations. The predicted masks exhibit sharper object contours, better-defined ground regions and cleaner obstacle separation. Furthermore, depth supervision enriches the encoder’s latent geometry, improving the delineation of traversable versus non-traversable areas. \rev{These findings indicate that dual-head supervision promotes semantically coherent and geometry-consistent spatial reasoning, enhancing perception robustness under diverse terrain.}

\textbf{Waypoint regression.} AIM-MT excels in waypoint prediction. However, Seq-DeepIPC narrows this gap as the input sequence length $K$ increases, benefiting from temporally fused latent embeddings that capture local motion continuity and long-term orientation cues. The GRU implicitly models short-horizon trajectory evolution, generating smoother waypoint transitions even without explicit attention or motion priors. Unlike other single-frame models, which rely solely on instantaneous appearance cues, Seq-DeepIPC leverages motion-consistent features across frames to reduce waypoint displacement variance. This advantage becomes pronounced in unstructured areas (stairs, grass, etc) where the robot’s orientation changes frequently and texture cues are sparse. Empirically, the waypoint MAE decreases monotonically with sequence length, confirming that temporal modeling provides an efficient approximation. These results suggest that Seq-DeepIPC’s temporal recurrence generalizes effectively to a natural mixed terrain.

\textbf{Control precision.} Both DeepIPC and Seq-DeepIPC outperform Huang in control MAE, demonstrating the importance of direct perception-to-control coupling mediated by explicit spatial grounding. Unlike the baseline which maps perspective features directly to actuation, Seq-DeepIPC’s joint perception branch links high-level semantic and depth features to low-level control dynamics, enabling context-aware motion responses. The geometry-informed latent representations enhance the model’s sensitivity to slope changes and surface irregularities, producing smoother velocity and orientation regulation. Temporal GRU fusion stabilizes these signals by filtering short-term fluctuations in perception output, reducing oscillations in the control loop. By aggregating features, the network learns to distinguish between transient camera shake and actual trajectory deviations. As a result, the robot exhibits fewer heading reversals and less jitter, particularly during transitions between different terrains where visual texture variance is high. Empirically, the control MAE reduction from $K{=}1$ to $K{=}3$ reflects more reliable policy consistency and reduced PID correction overhead. \rev{Overall, Seq-DeepIPC’s combination of temporal memory, geometry-aware supervision, and lightweight computation delivers robust control precision in navigating the legged robot in mixed terrains.}

\subsection{Qualitative Evaluation and Failure Analysis}\label{sec:qualitative}

The qualitative cases in Fig.~\ref{fig:qual_results} illustrate typical operational scenarios. In open environments as shown in Fig.~\ref{fig:qual_results}(a), the robot accurately distinguishes road and grass regions, with consistent segmentation and depth predictions over consecutive frames. The BEV map remains coherent, producing smooth waypoint trajectories and stable control commands. \rev{Furthermore, Fig.~\ref{fig:qual_results}(b) demonstrates the model's feasibility on uneven terrain. The robot successfully traverses a set of stairs embedded in a grassy hill. Although stairs violate the flat-ground assumption of the BEV projection (potentially causing geometric artifacts), the model successfully identifies the path. This robustness is attributed to the sequential fusion ($K=3$): as the legged robot climbs, its body pitch oscillates, causing significant camera jitter. The GRU integrates features over time, effectively smoothing these high-frequency disturbances and allowing the controller to treat the discontinuous stairs as a traversable slope.} In contrast, near tall structures as shown in Fig.~\ref{fig:qual_results}(c), GNSS errors distort the geodesic bearing used for coordinate transformation, yielding misplaced route points and misaligned control vectors. These results emphasize that the remaining bottleneck is not visual perception but external localization reliability. Such systematic errors can be mitigated by fusing GNSS with other robust sensors, applying temporal filtering of the absolute heading estimate, or adopting confidence-weighted waypoint sampling.

\rev{Online tests reveal that Seq-DeepIPC performs reliably in open areas and uneven terrain (stairs, grass)} However, its dependence on GNSS-only bearing estimation causes localization drift near tall buildings, where multipath interference corrupts coordinate transformation. The misalignment between global route points and local BEV grids leads to control offset. Importantly, perception quality remains intact, isolating the error source to global–local misprojection rather than network instability. Overall, Seq-DeepIPC demonstrates that \emph{temporal, geometric, and semantic integration} can jointly elevate end-to-end robot navigation performance. It establishes a practical framework for extending perception–control coupling from wheeled to legged robots that can traverse or navigate in a more diverse environment.

\section{Conclusion}

This paper presented Seq-DeepIPC, a sequential and multi-task end-to-end perception-to-control framework for legged robot navigation in mixed-terrain environments. Building on the previous work, DeepIPC, the model processes temporal RGB-D sequences through a lightweight encoder with dual-head perception for semantic segmentation and depth estimation. The outputs are projected into a bird’s-eye-view (BEV) map and fused by a GRU-based planner to produce geometry-aware control commands. Experiments show that temporal integration and depth supervision improve perception consistency, waypoint accuracy, and control stability. By coupling sensor fusion and control through learning, Seq-DeepIPC bridges the gap between intelligent sensing and autonomous decision-making.

The results highlight three core findings: (i) temporal recurrence enhances motion continuity and reduces perception noise, (ii) geometry-aware auxiliary supervision strengthens spatial reasoning, and (iii) balanced multi-task learning ensures stable convergence. Beyond quantitative gains, Seq-DeepIPC extends end-to-end navigation from wheeled to legged robot, demonstrating robust performance across road, stairs, and grass terrains. Although GNSS-only heading estimation remains sensitive near tall building and structures, the system performs reliably in open environments, suggesting the potential of integrating GNSS with other robust sensors for improved resilience.

\rev{To address the limitation of GNSS denial (e.g., the failure case in Fig.~\ref{fig:qual_results}(c)), future work will investigate fusing Visual Odometry (VO) or magnetometer-free VIO to maintain heading accuracy when satellite signals are obstructed. Moreover, other strategies such as adaptive temporal fusion, multi-modal sensing, and broader deployment on diverse robotic platforms can be explored to advance real-world autonomy}.


\bibliographystyle{IEEEtran}
\bibliography{references}

%\vspace{28mm}

\begin{IEEEbiography}[{\includegraphics[width=1in,height=1.25in,clip,keepaspectratio]{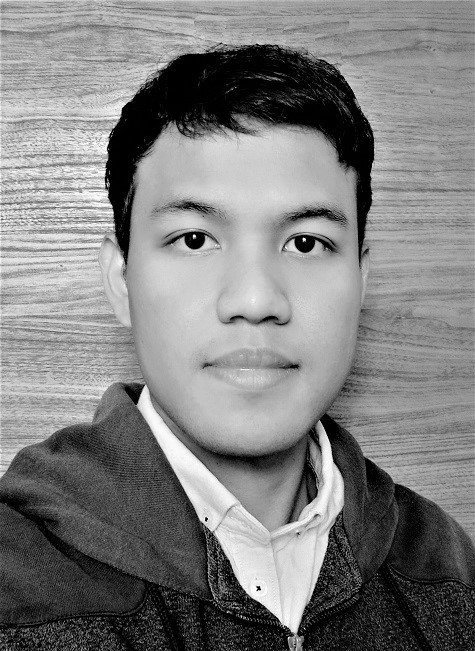}}]{Oskar Natan}
	(Member, IEEE) received his B.A.Sc. degree in Electronics Engineering and M.Eng. degree in Electrical Engineering from Politeknik Elektronika Negeri Surabaya, Indonesia, in 2017 and 2019, respectively. In 2023, he received his Ph.D.(Eng.) degree in Computer Science and Engineering from Toyohashi University of Technology, Japan. Since January 2020, he has been affiliated with the Department of Computer Science and Electronics, Universitas Gadjah Mada, Indonesia, first as a Lecturer and currently serves as an Assistant Professor. He has been serving as a reviewer/TPC member for some reputable journals and conferences. His research interests lie in the fields of deep learning, sensor fusion, hardware acceleration, and end-to-end systems. 
\end{IEEEbiography}


%\vspace{-8mm}

%\begin{IEEEbiography}{Jun Miura}
\begin{IEEEbiography}[{\includegraphics[width=1in,height=1.25in,clip,keepaspectratio]{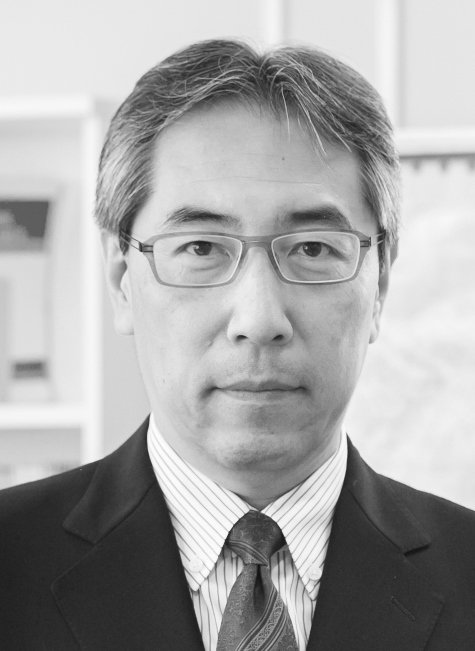}}]{Jun Miura}
	(Member, IEEE) received his B.Eng. degree in Mechanical Engineering and his M.Eng. and Dr.Eng. degrees in Information Engineering from the University of Tokyo, Japan, in 1984, 1986, and 1989, respectively. From 1989 to 2007, he was with the Department of Computer-controlled Mechanical Systems, Osaka University, Japan, first as a Research Associate and later as an Associate Professor. From March 1994 to February 1995, he served as a Visiting Scientist at the Department of Computer Science, Carnegie Mellon University, USA. In 2007, he became a Professor at the Department of Computer Science and Engineering, Toyohashi University of Technology, Japan, where he remains to the present. To date, he has received plenty of awards and authored or co-authored more than 265 peer-reviewed scientific articles in the field of robotics and autonomous systems in internationally reputable journals and conferences.
\end{IEEEbiography}

\vfill
%\newpage

\end{document}